\documentclass[10pt,twocolumn,letterpaper]{article}

\usepackage{cvpr}              
\usepackage{bm} 
\usepackage{multirow}
\usepackage{colortbl}

\definecolor{cvprblue}{rgb}{0.21,0.49,0.74}
\usepackage[pagebackref,breaklinks,colorlinks,allcolors=cvprblue]{hyperref}

\def\paperID{5933} 
\def\confName{CVPR}
\def\confYear{2025}

\title{BIP3D: Bridging 2D Images and 3D Perception for Embodied Intelligence}

\author{Xuewu Lin, Tianwei Lin, Lichao Huang, Hongyu Xie, Zhizhong Su\\
Horizon Robotics, Beijing, China\\
{\tt\small xuewu.lin@horizon.auto}
}

\begin{document}

\maketitle
\begin{abstract}
In embodied intelligence systems, a key component is 3D perception algorithm, which enables agents to understand their surrounding environments.
Previous algorithms primarily rely on point cloud, which, despite offering precise geometric information, still constrain perception performance due to inherent sparsity, noise, and data scarcity.
In this work, we introduce a novel image-centric 3D perception model, BIP3D, which leverages expressive image features with explicit 3D position encoding to overcome the limitations of point-centric methods.
%
Specifically, we leverage pre-trained 2D vision foundation models to enhance semantic understanding, and introduce a spatial enhancer module to improve spatial understanding. Together, these modules enable BIP3D to achieve multi-view, multi-modal feature fusion and end-to-end 3D perception.
%
In our experiments, BIP3D outperforms current state-of-the-art results on the EmbodiedScan benchmark, achieving improvements of 5.69\% in the 3D detection task and 15.25\% in the 3D visual grounding task. 
Code will be released at \url{https://github.com/HorizonRobotics/BIP3D}.


%
\end{abstract}

\begin{table*}
  \centering
  \begin{tabular}{@{}l|ccccccc@{}}
    \toprule
    Methods & Main Modality &  Point/Depth & Image & Multi-view Fusion & Text & Single-stage & Det\&Grounding\\
    \midrule
    ~\cite{votenet,liu2021groupfree,rukhovich2022fcaf3d,kolodiazhnyi2024unidet3d} & Point & \checkmark &  &  &  &  &  \\
    ~\cite{qi2020imvotenet,xu2023nerfdet} & Image &  & \checkmark & \checkmark &  &  & \\
    ~\cite{piekenbrinck2024rgb}& Image & \checkmark & \checkmark &  &  &  & \\
    ~\cite{achlioptas2020referit3d,chen2020scanrefer,roh2022languagerefer,yang2021sat} & Point & \checkmark &  &  & \checkmark &  &\\
    ~\cite{miyanishi2024cross3dvg}& Point & \checkmark & \checkmark & \checkmark & \checkmark &  &\\
    ~\cite{butd-detr,3DOGSFormer,3dsps} & Point & \checkmark &  &  & \checkmark & \checkmark & \\
    ~\cite{wang2024embodiedscan,zhengdenseg} & Point & \checkmark & \checkmark & \checkmark & \checkmark & \checkmark & \\
    \midrule
    BIP3D & Image & \checkmark & \checkmark & \checkmark & \checkmark & \checkmark & \checkmark \\
    \bottomrule
  \end{tabular}
  \caption{Comparison of Key Elements among Different Methods. "Single-stage" indicates whether the model is a single-stage visual grounding approach, "Det\&Grounding" denotes whether the model can simultaneously perform 3D detection and grounding.}
  \label{tab:related_works}
\end{table*}
\section{Introduction}
\label{sec:intro}
\begin{figure}
    \centering
    \includegraphics[width=0.99\linewidth]{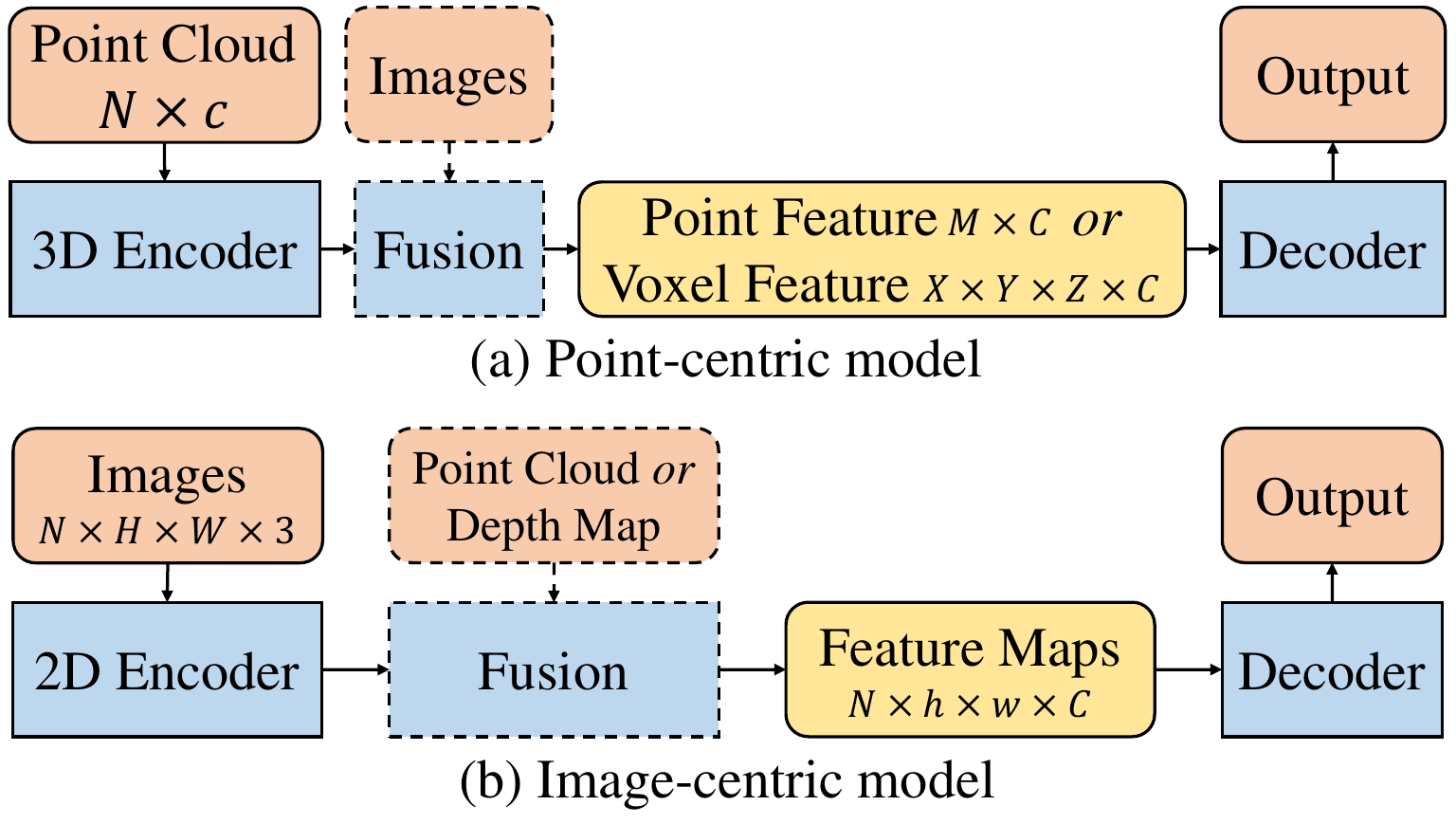}
    \caption{Comparison of Point-centric and Image-centric Model Architectures. \emph{Dashed boxes denote optional pluggable modules.} (a) The point-centric model centers its parameters within the 3D encoder, utilizing feature representations like point or 3D voxel features. (b) By contrast, the image-centric model emphasizes the 2D encoder, using 2D feature maps for its representations.}
    \label{fig:pcic}
\end{figure}

3D perception models are utilized to estimate the 3D pose, shape, and category of objects of interest in a scene, typically outputting 3D bounding boxes or segmentation masks. In the field of embodied intelligence, these models generally serve to provide essential input for planning modules or serve as crucial algorithmic components in cloud-based data systems. Enhancing the accuracy of 3D perception holds significant research value. 
As shown in Figure~\ref{fig:pcic}(a), current mainstream 3D perception models extract features from the point cloud (using PointNet++~\cite{qi2017pointnet++} or 3D CNN~\cite{choy20194dconv}) and generate perception results based on these point features. 
%

While point clouds offer precise geometric information that has advanced 3D perception, several challenges remain~\cite{depthcam}: 
(1) High-quality point clouds are difficult to obtain, because depth sensors often have limitations,  such as struggling with reflective or transparent objects, long distances, or intense lighting conditions.
(2) Point clouds are sparse, lack texture, and can be noisy, leading to detection errors and a limited performance ceiling.
(3) Collecting and annotating point cloud data is costly, which poses a challenge for acquiring large-scale training data.

These challenges limit the performance of point-centric models. 
In contrast, the abundance of image data has accelerated advancements in vision models, with 2D vision foundation models exhibiting strong semantic understanding and generalization capabilities. 
Therefore, transferring 2D vision foundation models (e.g. CLIP~\cite{clip, zhong2022regionclip}, EVA~\cite{fang2024eva} and DINO~\cite{oquab2023dinov2}) 
to the 3D domain holds significant potential for enhancing 3D task performance.

In this paper, we propose BIP3D, an image-centric 3D perception model (Figure~\ref{fig:pcic}(b)) that performs 3D object detection and 3D visual grounding by fusing multi-view image and text features, and can accept depth maps as auxiliary inputs for enhanced precision.
Our model is based on the 2D model, GroundingDINO~\cite{groundingdino}, sharing a similar overall network architecture and initialized with its model weights, thereby inheriting the strong generalization capabilities of GroundingDINO. Unlike GroundingDINO, our model can accept an arbitrary number of posed images as input and output 3D bounding boxes.
The main improvements include the following three aspects:
\textbf{(1) Camera Modeling:} Explicitly constructing the camera model, which supports the input of intrinsic and extrinsic camera parameters, providing 3D position encoding for 2D image features and relative position information between 3D objects and images;
\textbf{(2) Multi-View Fusion:} Modifying the 2D deformable attention in the DINO decoder to a 3D form, achieving dynamic multi-view feature fusion;
\textbf{(3) Multi-Modal Fusion:} Adding a depth map encoding branch to realize multi-modal feature fusion between images and depth maps, enhancing 3D perception performance.

Unlike existing mainstream approaches, we focus more on image feature encoding, with the primary layers dedicated to processing 2D image features rather than 3D point features. The feature representation is consistently organized as structured multi-view, multi-scale feature maps. Compared to point clouds, image features have higher information density, higher signal-to-noise ratios, and better scene generalization. Moreover, our model supports RGB-only input, allowing for the rapid collection of large amounts of data for embodied intelligence scenarios through crowd-sourcing, which is crucial for continuously improving model performance.

We conduct extensive experiments on the EmbodiedScan benchmark~\cite{wang2024embodiedscan}, which includes data from the ScanNet~\cite{dai2017scannet}, 3RScan~\cite{3rscan}, and Matterport3D~\cite{chang2017matterport3d} datasets, all of which have been more thoroughly annotated. Compared to other datasets, this benchmark is better suited for evaluating the generalization performance of 3D models.
Firstly, we complete the 3D detection task using category grounding and achieved state-of-the-art (SOTA) performance on the 3D detection benchmark. Compared to the EmbodiedScan baseline, the primary metric (AP$_{3D}$@0.25) improved by 5.69\%.
Secondly, we also achieve SOTA performance in the 3D visual grounding task, significantly outperforming existing methods. On the validation set, we improved by 15.25\% over the EmbodiedScan baseline, and on the test set, we surpassed the CVPR 2024 Challenge 1st place, DenseG~\cite{zhengdenseg}, by 2.49\%.
Additionally, we perform ablation studies to validate the effectiveness of our key improvements.
In summary, our main contributions are:

\begin{itemize}
    \item An image-centric 3D perception model, BIP3D, is proposed, which takes multi-view images and text as input and outputs 3D perception results.
    \item State-of-the-art performance is achieved on the EmbodiedScan 3D detection and grounding benchmark.
    \item Deep analyses of the improvements are conducted to provide guidance for future image-centric models.
\end{itemize}

\section{Related Works}
\label{sec:related works}

\subsection{3D Object Detection}
 3D detection algorithms are primarily categorized into outdoor and indoor scene applications. Due to notable differences in perception range, target types, and sensor types, the development of algorithms for these two settings varies considerably. This paper focuses on the indoor scenes.

For most indoor scenes, depth cameras provide accurate 3D point clouds, which can be directly aggregated across multiple frames for early-fusion. Therefore, most existing methods focus on point clouds.
VoteNet~\cite{votenet} uses PointNet++~\cite{qi2017pointnet++} to extract point cloud features and aggregates them using Hough voting.
Group-Free~\cite{liu2021groupfree} leverages a Transformer decoder to implicitly aggregate features from multiple points, reducing errors from hand-crafted grouping.
FCAF3D~\cite{rukhovich2022fcaf3d} voxelizes the point cloud, extracts features using a U-Net architecture with sparse 3D convolutions, and outputs boxes through convolutional decoder.
UniDet3D~\cite{kolodiazhnyi2024unidet3d} uses 3D sparse convolutions and a transformer decoder for 3D detection.
ImVoteNet~\cite{qi2020imvotenet} enhances VoteNet by incorporating RGB image features for voting.
EmbodiedScan~\cite{wang2024embodiedscan} projects sparse voxels onto multi-view images to sample features, achieving multi-modal feature fusion.
ImVoxelNet~\cite{rukhovich2022imvoxelnet} and NeRF-Det~\cite{xu2023nerfdet} use only images as input, convert 2D features to 3D voxel features via inverse perspective mapping, and output dense prediction boxes through 3D convolutions.
RGB-D Cube R-CNN~\cite{piekenbrinck2024rgb} also fuses multi-modal features but is limited to single-view input.
In contrast, our BIP3D is image-centric and achieves multi-view and multi-modal feature fusion. Additionally, we perform 3D detection in the category grounding manner, which makes BIP3D extendable to an open-set detector.

\subsection{3D Visual Grounding}
3D visual grounding involves generating the 3D bounding box for a target in the environment based on a text instruction.
Early methods divide this task into two stages: the first stage uses ground truth or a pre-trained 3D detector to generate object proposals, while the second stage scores each proposal using text features, selecting those that exceed a threshold as final results.
ReferIt3DNet~\cite{achlioptas2020referit3d} and ScanRefer~\cite{chen2020scanrefer} introduce two 3D visual grounding datasets and propose point-centric two-stage models as baselines. LanguageRefer~\cite{roh2022languagerefer} encodes both proposals and text as inputs to a language model, using it to assess proposal confidence. SAT~\cite{yang2021sat} highlights the issues of sparse, noisy, and limited semantic information in point clouds and incorporates images into training to enhance model performance. Cross3DVG~\cite{miyanishi2024cross3dvg} leverages CLIP~\cite{clip} features from multi-view images to boost grounding effects.

In addition to these two-stage models, single-stage grounding approaches have emerged in the 3D domain, inspired by 2D grounding techniques~\cite{kamath2021mdetr}.
These methods directly fuse text and scene features and produce grounding results. BUTD-DETR~\cite{butd-detr} achieves single-stage 3D grounding with a point encoder and transformer decoder structure, offering the option to add extra proposals to improve performance.
3DSPS~\cite{3dsps} transforms 3D grounding into a keypoint selection task, eliminating the need to separate detection and scoring.
3DOGSFormer~\cite{3DOGSFormer} enables simultaneous processing of multiple text instructions.
%
Table~\ref{tab:related_works} highlights  the differences between our BIP3D and existing 3D detection and grounding methods.

\subsection{Vision Foundation Model}
A foundation model is one that has been pre-trained on large datasets and can achieve good performance on multiple downstream tasks.

In the field of 2D vision, foundation models are categorized by their pre-training approaches: supervised training~\cite{vit22b}, unsupervised training~\cite{mae,bao2021beit}, image contrastive learning~\cite{mocov3,oquab2023dinov2}, and image-text contrastive learning~\cite{clip}.
Foundation models trained with image-text contrastive learning exhibit the strongest generalization and zero-shot capabilities. These include methods that align full images with text~\cite{clip,evaclip} and those that align cropped images with text~\cite{zhong2022regionclip,glip,groundingdino}. Aligning cropped images with text not only improves classification but also enhances localization, which is crucial for grounding tasks.

Building 3D foundation models typically follows two technical routes:
(1) Contrastive learning with 3D point clouds and text features~\cite{zhang2022pointclip,huang2023clip2point,zhou2023uni3d}. This approach requires extensive 3D data, which is less abundant than 2D data. Most available data is 3D object-text, with limited 3D scene-text data, leading to poor performance in 3D detection and grounding tasks.
(2) Extracting image features using 2D foundation models and obtaining 3D features through dense mapping~\cite{jatavallabhula2023conceptfusion,3dllm}. These methods require depth information, usually from depth sensors or SLAM.

This paper aims to enhance 3D perception by leveraging existing 2D foundation models.
We choose GroundingDINO~\cite{groundingdino} as the base model and avoid dense mapping, directly using multi-view feature maps for 3D detection and grounding.

\section{Method}
\begin{figure*}
    \centering
    \includegraphics[width=0.99\linewidth]{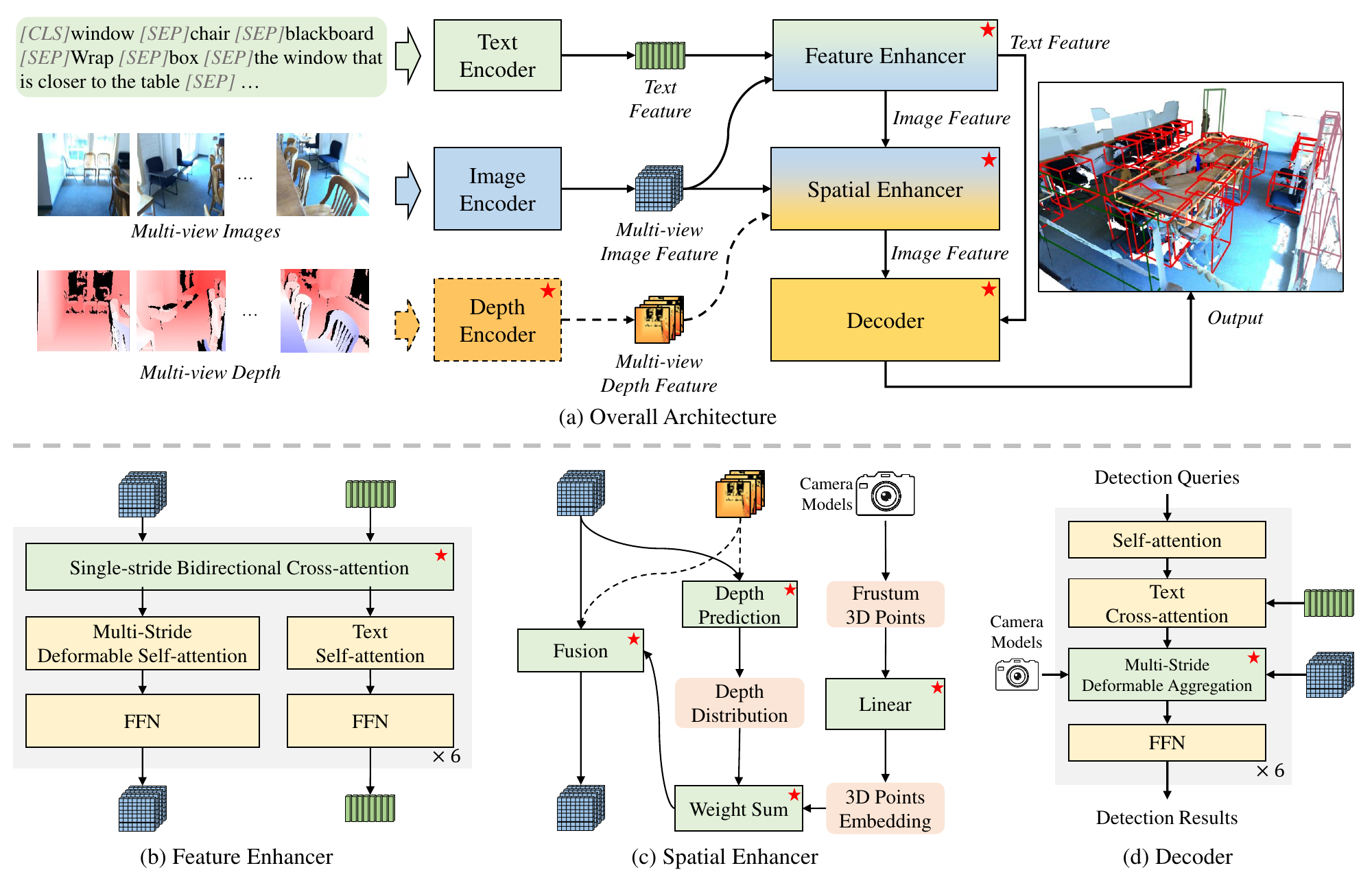}
    \caption{The Architecture Diagram of BIP3D, where $\textcolor{red}{\star}$ indicates the parts that have been modified or added compared to the base model, GroundingDINO~\cite{groundingdino}, and dashed lines indicate optional elements.}
    \label{fig:structure}
\end{figure*}

In Figure~\ref{fig:structure}(a), we demonstrate the overall network architecture of BIP3D, which takes multi-view images, text instructions and optional  depth maps as input.
The output is 3D bounding boxes for the objects specified by the text, with support for multiple text instructions that may correspond to multiple objects.
The network comprises six main modules:
text, image, and depth encoders that individually process the input components into high-dimensional features;
a feature enhancer module that fuses the text and image features (Sec.~\ref{sec:feature_enhancer});
a spatial enhancer module that performs 3D position encoding and depth fusion, enriching the image features with 3D cues (Sec.~\ref{sec:pe-depthfusion}); 
finally, a transformer decoder that generates the 3D bounding box from the multi-view images and text features (Sec.~\ref{sec:decoder}).

\subsection{Feature Enhancer}
\label{sec:feature_enhancer}
Since our model needs to support multi-view inputs, the number of image feature vectors is $N\sum(HW/s_i^2)$, where $N$ is the number of views. Under the setting of EmbodiedScan, $N$ equals 50, which makes the computation and memory consumption of cross-attention excessively high to be feasible.
Therefore, we only use the feature map with the maximum stride for cross-attention. For the image features in other strides, text information is indirectly obtained through intra-view multi-stride deformable attention.

\subsection{Spatial Enhancer}
\label{sec:pe-depthfusion}
Image features only contain information from the current camera coordinate system and are insensitive to camera models, particularly the extrinsic parameters. Therefore, we explicitly perform 3D encoding of camera models. Specifically, given a camera model as a projection function $\operatorname{C}$:
\[
[x, y, z] = \operatorname{C}([u, v, d])
\]
\[
[u, v] = \operatorname{C}^{-1}([x, y, z])
\]
where $[x,y,z]$ represents the 3D coordinates in the perception coordinate system, $[u,v]$ represents the 2D pixel coordinates, and $d$ is the depth in the camera coordinate system. For example, in a common pinhole camera model:
\[
[x, y, z, 1]^{\mathrm{T}} = \bm{\mathrm{T}}_{E}\bm{\mathrm{T}}_{I}[ud, vd, d, 1]^{\mathrm{T}}
\]
where $\bm{\mathrm{T}}_{E} \in \mathbb{R}^{4\times 4}$ is the extrinsic matrix and $\bm{\mathrm{T}}_{I} \in \mathbb{R}^{4\times 4}$ is the intrinsic matrix. Based on the camera model 
$\operatorname{C}$, we uniformly sample some 3D points within its view frustum and project these points into the perception coordinate system:
\[
\left\{\operatorname{C}\left(u_i, v_j, \frac{k\cdot D}{K}\right) \mid 0 \leq i < w, 0 \leq j < h, 0 \leq k < K\right\}
\]
Here, $D$ and $K$ are hyperparameters, representing the maximum depth and the number of sampling points, respectively. We have omitted the stride dimension for simplicity. Then, we use a linear layer to transform the 3D point coordinates into high-dimensional features, resulting in a series of point position embedding:
\[
\bm{\mathrm{PPE}}_{i,j,k}=\operatorname{Linear}_1\left(\operatorname{C}\left(u_i, v_j, \frac{k\cdot D}{K}\right)\right)
\]

We predict the depth distribution $\bm{\mathrm{DT}}$ using the image feature and depth feature, and use this distribution to weight the point embedding to obtain image position embedding:
\[
\bm{\mathrm{DT}}_{i,j} = \operatorname{Linear}_2(\operatorname{Fusion_3}(\bm{\mathrm{I}}_{i,j}, \bm{\mathrm{D}}_{i,j})) \in \mathbb{R^{K}}
\]
\[
\bm{\mathrm{IPE}}_{i,j} = \sum_{k}(\bm{\mathrm{PPE}}_{i,j,k}\times \bm{\mathrm{DT}}_{i,j,k})
\]

Finally, we get the updated image feature $\bm{\mathrm{I}}'$ by fuse the image feature $\bm{\mathrm{I}}$, depth feature $\bm{\mathrm{D}}$, and image position embedding $\bm{\mathrm{IPE}}$:
\[
\bm{\mathrm{I}}_{i,j}' = \operatorname{Fusion}_4(\bm{\mathrm{I}}_{i,j}, \bm{\mathrm{D}}_{i,j}, \bm{\mathrm{IPE}}_{i,j})
\]
where 
\[\operatorname{Fusion}_i(x_1,x_2,...)=\operatorname{Linear}_i(\operatorname{Concat}(x_1, x_2,...))
\]

It is worth noting that, while PETR~\cite{liu2022petr} also designs 3D position embeddings, it does not consider depth distribution $\bm{\mathrm{DT}}$. As a result, its position embeddings are independent of image features, which limits their capabilities.

\subsection{Decoder with Multi-view Fusion}
\label{sec:decoder}
The decoder of GroundingDINO uses 2D deformable attention~\cite{deformabledetr} to achieve feature interaction between image and queries, which is insufficient for 3D multi-view applications. In this paper, we replace it with 3D deformable aggregation~\cite{lin2022sparse4d}, and conduct certain adaptations and improvements. Specifically, for each query, we maintain a corresponding 3D bounding box, represented as:
\[
\bm{\mathrm{B}}=\left[x,y,z,l,w,h,roll,pitch,yaw\right]
\]

We sample $M$ 3D key points within the 3D bounding box: (1) regress a series of offsets $O_{l}$ based on the query feature to obtain 3D learnable key points, and (2) set some fixed offsets $O_{fix}$ based on prior knowledge of 3D detection to get fix key points, such as the stereo center and the centers of the six faces of the bounding box.
\[O_{3d} = O_{l} \cup O_{fix} = \left\{\Delta[x,y,z]_i | 1\leq i\leq M\right\}\]
\[
P_{3D} = \left\{(\Delta[x,y,z]_i * \bm{\mathrm{R}}^\mathrm{T} )*[l, w, h]+ [x,y,z] | 1\leq i\leq M\right\}
\]
Here, $\bm{\mathrm{R}}$ is the rotation matrix derived from the Euler angles $[roll,pitch,yaw]$, We project $P_{3D}$ onto the multi-view feature maps using the camera model to obtain $P_{2D}$, and perform feature sampling.
\[
P_{2D}=\left\{\operatorname{C}^{-1}_{j}(P_{3D,i})| 1\leq i\leq M, 1\leq j\leq N\right\}
\]
\[
\bm{\mathrm{F}}=\left\{\operatorname{Bilinear}(\bm{\mathrm{I}}, P_{2D,i,j}) | 1\leq i\leq M,1\leq j\leq N\right\}
\]

Finally, we combine the query feature, 3D bounding box, and camera parameters to predict the weighting coefficients, which are used to obtain the updated query, thus completing the feature transfer from image features to the query.
\[
\bm{\mathrm{W}}=\operatorname{Softmax}(\operatorname{Linear}(\operatorname{Fusion}(\bm{\mathrm{Q}},\bm{\mathrm{B}},\operatorname{C})))\in \mathbb{R^{M\times N}}
\]
\[\bm{\mathrm{Q}}'=\sum_{i,j}{\bm{\mathrm{W}}_{i,j}\bm{\mathrm{F}}_{i,j}}\]
Unlike in autonomous driving scenarios~\cite{lin2022sparse4d}, in embodied intelligence scenarios, the number of views and the extrinsic parameters of cameras are not fixed and are constantly changing. This results in significant variations in invalid sampling points (those outside the view frustum), making the model's convergence more challenging. Therefore, we perform explicit filtering and set the weights $\bm{\mathrm{W}}$ of invalid key points to zero.

\subsection{Camera Intrinsic Standardization}
We found that the generalization of camera parameters in image-centric models is relatively poor, especially when (1) the richness of camera parameters in the dataset is insufficient or (2) there is no depth or point cloud to provide geometric information. To mitigate this problem, we propose a method for camera intrinsic standardization.
Specifically, given an image and its camera intrinsics $\operatorname{C}_{I}$, and a predefined standardized camera intrinsics $\operatorname{C}_{I}'$, we transform the image to a virtual camera coordinate system by inverse projection $\operatorname{C}_I(\operatorname{C}_I'^{-1}(i,j))$,  where $(i,j)$ are the pixel coordinates.
For a standard pinhole camera, the inverse projection formula can be reduced to an affine transformation. We use the mean of the intrinsic parameters from the training set as $\operatorname{C}_{I}'$. During both training and inference, we standardize the intrinsics of all input images and use the transformed images as inputs to the model.

\begin{table*}[ht]
  \centering
  \begin{tabular}{@{}l|c|ccc|ccc|ccc@{}}
    \toprule
    Methods & Overall & Head & Common & Tail & Small & Medium & Large & ScanNet & 3RScan & MP3D\\
    \midrule
    VoteNet~\cite{votenet} & 5.18   & 10.87  & 2.41  & 2.07  & 0.16  & 5.30  & 5.99  & 9.90 & 7.69 & 3.82 \\
    ImVoxelNet~\cite{rukhovich2022imvoxelnet} & 8.08 & 3.11 & 7.05 & 3.73 & 0.06 & 7.95 & 9.02 & 11.91 & 2.17 & 5.24 \\
    FCAF3D~\cite{rukhovich2022fcaf3d} & 13.86 & 22.89 & 9.61 & 8.75 & 2.90 & 13.90 & 10.91 & 21.35 & 17.02 & 9.78 \\
    EmbodiedScan-D~\cite{wang2024embodiedscan} & 15.22 & \underline{24.95} & 10.81 & 9.48 & 3.28 & 15.24 & 10.95 & \underline{22.66} & 18.25 & \textbf{10.91}\\
    \rowcolor{gray!20} BIP3D-RGB & \underline{17.40} & 24.20 & \underline{14.76} & \underline{12.94} & \underline{3.45} & \underline{18.19} & \underline{14.06} & 20.38 & \underline{27.23} & 8.77  \\
    \rowcolor{gray!20} BIP3D & \textbf{20.91} & \textbf{27.57} & \textbf{18.77} & \textbf{16.03} & \textbf{5.72} & \textbf{21.48} & \textbf{15.20} & \textbf{23.47} & \textbf{32.48} & \underline{10.09}  \\
    \bottomrule
  \end{tabular}
  \caption{Experiment Results of 3D Detection Task on the EmbodiedScan Validation Dataset. `EmbodiedScan-D' denotes the detection model provided by EmbodiedScan, and `MP3D' stands for Matterport3D.}
  \label{tab:detection3d-main}
\end{table*}
\begin{table*}[t]
  \centering
  \begin{tabular}{@{}l|l|c|cc|cc|ccc@{}}
    \toprule
    Eval Set& Methods & Overall & Easy & Hard & View-dep & View-indep  & ScanNet & 3RScan & MP3D\\
    \midrule
    \multirow{3}{*}{Validation} & EmbodiedScan~\cite{wang2024embodiedscan} & 39.41 & 40.12 & 31.45 & 40.21 & 38.96 & 41.99 & 41.53 & 30.29  \\
    & \cellcolor{gray!20}BIP3D-RGB
    & \cellcolor{gray!20}46.04
    & \cellcolor{gray!20}46.25
    & \cellcolor{gray!20}43.76
    & \cellcolor{gray!20}46.16
    & \cellcolor{gray!20}45.98
    & \cellcolor{gray!20}51.82
    & \cellcolor{gray!20}46.30
    & \cellcolor{gray!20}33.27 \\
    & \cellcolor{gray!20}BIP3D
    & \cellcolor{gray!20}\textbf{54.66}
    & \cellcolor{gray!20}\textbf{55.07}
    & \cellcolor{gray!20}\textbf{50.12}
    & \cellcolor{gray!20}\textbf{55.78}
    & \cellcolor{gray!20}\textbf{54.03}
    & \cellcolor{gray!20}\textbf{61.23}
    & \cellcolor{gray!20}\textbf{55.41}
    & \cellcolor{gray!20}\textbf{39.36} \\
    \midrule
    \multirow{5}{*}{Test} & EmbodiedScan~\cite{wang2024embodiedscan} &39.67 & 40.52 & 30.24 & 39.05 &39.94&- & - & - \\
    & SAG3D*~\cite{liangspatioawaregrouding3d} & 46.92 & 47.72 & 38.03 & 46.31 & 47.18 & - & - & -\\
    & DenseG*~\cite{zhengdenseg} & 59.59 & 60.39 & 50.81 & 60.50 & 59.20 & - & - & -\\
    & \cellcolor{gray!20}BIP3D 
    & \cellcolor{gray!20}57.05
    & \cellcolor{gray!20}57.88
    & \cellcolor{gray!20}47.80
    & \cellcolor{gray!20}57.16
    & \cellcolor{gray!20}57.00
    & \cellcolor{gray!20}- 
    & \cellcolor{gray!20}- 
    & \cellcolor{gray!20}- \\
    & \cellcolor{gray!20}BIP3D*
    & \cellcolor{gray!20}\textbf{62.08}
    & \cellcolor{gray!20}\textbf{63.08}
    & \cellcolor{gray!20}\textbf{50.99}
    & \cellcolor{gray!20}\textbf{61.84}
    & \cellcolor{gray!20}\textbf{62.18}
    & \cellcolor{gray!20}- 
    & \cellcolor{gray!20}- 
    & \cellcolor{gray!20}- \\
    \bottomrule
    
  \end{tabular}
  \caption{Experiment Results of 3D Visual Grounding Task on the EmbodiedScan Dataset. `View-dep' and `View-indep' refer to view-dependent and view-independent, respectively. `*' denotes model ensemble.}
  \label{tab:grounding-main}
\end{table*}
\subsection{Training}
We apply a one-to-one matching loss~\cite{DETR} to the output of each decoder layer. The loss consists of three components:
\[
Loss=\lambda_1 L_{cls}+\lambda_2 L_{center}+\lambda_3 L_{box}
\]
where 
$L_{cls}$ is the contrastive loss between queries and text features for classification, using Focal Loss; 
$L_{center}$ is the center point regression loss, using L2 Loss;
$L_{box}$ is the bounding box regression loss for 9-DoF detection.
To avoid ambiguities in $[w,l,h]$ and $[roll,pitch,yaw]$ due to insufficient definition of object orientation, we introduce a simplified Wasserstein distance as $L_{box}$. Specifically, for a bounding box, we assign it a 3D Gaussian distribution $N(\mu,\bm{\mathrm{\Sigma}}^2)=N([x,y,z],\bm{\mathrm{R}} \bm{\mathrm{S}}^2 \bm{\mathrm{R}}^\mathrm{T})$ where $\bm{\mathrm{S}}$ is a diagonal matrix with $[w,l,h]$ along its diagonal. Given $\bm{\mathrm{B}}_{gt}$ and $\bm{\mathrm{B}}_{pred}$, the formula for $L_{box}$ is defined as follows:
\[
L_{box} = \sqrt{||\mu_{gt}-\mu_{pred}||_2 + ||\bm{\mathrm{\Sigma}}_{gt}-\bm{\mathrm{\Sigma}}_{pred}||_F}
\]

Since the rotation matrix $\bm{\mathrm{R}}$ is orthogonal, $\bm{\mathrm{\Sigma}}$ equals $\bm{\mathrm{R}} \bm{\mathrm{S}} \bm{\mathrm{R}}^\mathrm{T}$. Follow DINO~\cite{zhangdino}, we also incorporate a denoising task to assist in training.

For 3D detection, we implement it in the form of category grounding. During training, we sample a subset of categories and set the text to \emph{“[CLS]cls\_1[SEP]cls\_2...[SEP]”}.
After training the 3D detection model, we load its weights as pretraining weights and then train the referring grounding model.
During the referring grounding training, for each training sample, we randomly sample 0 to $x$ descriptions and set the text to \emph{“[CLS]exp\_1[SEP]exp\_2...[SEP]”}.

\section{Experiments}

\begin{table}
  \centering
  \begin{tabular}{@{}l|ccccc@{}}
    \toprule
    & \multicolumn{5}{c}{Modules' Parameters (MB)} \\
    Model & \scalebox{1.5}{$\operatorname{\epsilon}$}$_{2D}$ & \scalebox{1.5}{$\operatorname{\epsilon}$}$_{3D}$ & \scalebox{1.5}{$\operatorname{\epsilon}$}$_{text}$ & Decoder & Others\\
    \midrule
    ES-G~\cite{wang2024embodiedscan} & 1.43 & 87.31 & 119.06 & 11.96 & 0.00\\
    BIP3D & 28.29 & 0.09 & 104.03 & 12.82 & 21.61 \\
    \bottomrule
  \end{tabular}
  \caption{Comparison of Modules' Parameters. `ES-G' denotes the grounding model in EmbodiedScan, and `\scalebox{1.5}{$\operatorname{\epsilon}$}' denotes encoder.}
  \label{tab:parameters}
\end{table}

\begin{table*}
  \centering
  \begin{tabular}{@{}l|c|c|c|ccc|ccc@{}}
    \toprule
    Methods & Pretrain & Inputs & Overall & Head & Common & Tail & Small & Medium & Large\\
    \midrule
    \multirow{2}{*}{EmbodiedScan~\cite{wang2024embodiedscan}} & ImageNet & RGB-D & 15.22 & 24.95 & 10.81 & 9.48 & 3.28 & 15.24 & 10.95 \\
     & GroundingDINO & RGB-D & 16.33 & 26.58 & 12.26 & 9.70 & 3.55 & 16.59 & 12.97 \\
    \midrule
    \multirow{4}{*}{BIP3D} & ImageNet  & RGB & 11.74 & 16.66 & 10.68 & 7.54 & 2.29 & 12.46 & 9.98 \\
     & GroundingDINO  & RGB & 17.40 & 24.20 & 14.76 & 12.94 & 3.45 & 18.19 & 14.06\\
     & ImageNet  & RGB-D & 14.92 & 20.93 & 12.63 & 10.93 & 3.64 & 15.99 & 10.29 \\
     & GroundingDINO  & RGB-D & 20.91 & 27.57 & 18.77 & 16.03 & 5.72 & 21.48 & 15.20\\
    \bottomrule
  \end{tabular}
  \caption{Ablation Study on 2D Pretrained Weights for 3D Detection.}
  \label{tab:ablation pretrain}
\end{table*}

\begin{table*}
  \centering
  \begin{tabular}{@{}c|c|ccc|c|ccc|ccc@{}}
    \toprule
    & & \multicolumn{3}{c|}{Train Set} & & \multicolumn{3}{c|}{Eval Set}& \multicolumn{3}{c}{Eval Category Split} \\
    Inputs & CIS & ScanNet & 3RScan & MP3D &  Overall & ScanNet & 3RScan & MP3D & Head & Common & Tail\\
    \midrule
    \multirow{6}{*}{RGB} & & \checkmark & & & 6.08 & 16.25 & 2.70 & 3.14 & 11.86  &4.53 & 1.50  \\
    & \checkmark & \checkmark & & & 6.88 & 17.73 & 4.54 & 3.98  & 12.89 & 5.57  &1.78\\
    &  & \checkmark & \checkmark & & 14.63 & 18.20 & 23.94 & 4.42 & 19.62 & 13.36 & 10.59  \\
    & \checkmark & \checkmark & \checkmark & & 15.53 & 18.88 & 26.88 & 4.74 & 21.68 & 13.09 & 11.56  \\
    &  & \checkmark & \checkmark  & \checkmark & 16.67 & 18.90  &25.42 &  8.90 & 22.21 & 14.48 & 13.07  \\
    & \checkmark & \checkmark & \checkmark  & \checkmark& 17.40 & 20.38 & 27.23 & 8.77 & 24.20 & 14.76 & 12.94  \\
    \midrule
    \multirow{6}{*}{RGB-D} & & \checkmark & & & 7.82 & 20.60 & 3.05 & 3.94 & 14.44 & 6.38 & 2.18  \\
    &\checkmark & \checkmark & & & 8.59 & 22.29 & 4.27 & 5.78 & 15.12 & 7.71 & 2.42  \\
    &  & \checkmark & \checkmark & & 17.80 & 22.79 & 30.04 & 4.92 & 23.24 & 16.59 & 13.21  \\
    & \checkmark & \checkmark & \checkmark & & 19.16 & 23.35 & 33.11 & 4.95 & 25.26 & 17.42 & 14.43  \\
    &  & \checkmark & \checkmark  & \checkmark & 19.78 & 22.01 & 30.95 & 10.41 & 26.74 & 16.43 & 15.95  \\
     & \checkmark & \checkmark & \checkmark  & \checkmark & 20.91 & 23.47 & 32.48 & 10.09 & 27.57 & 18.77 & 16.03  \\
    \bottomrule
  \end{tabular}
  \caption{Ablation Study on the Impact of Camera Intrinsic Standardization on 3D Detection.}
  \label{tab:ablation cis}
\end{table*}

\begin{table*}
  \centering
  \begin{tabular}{@{}l|c|ccc|ccc|ccc@{}}
  \toprule
  Loss Type & Overall & Head & Common & Tail & Small & Medium & Large& ScanNet & 3RScan & MP3D\\
  \midrule
  CCD & 0.64 & 0.32 & 1.15 & 0.41 & 0.96 & 0.27  &0.00 & 1.19 & 1.37 & 0.12 \\
    L1 & 17.79 & 24.62 & 15.73 & 12.62 & 3.86 & 19.14 & 14.40 & 21.90 & 28.21 & 8.92  \\
    PCD & 20.21 & \textbf{27.62} & 17.71 & 14.91 & 5.47 & 20.94 & 14.30 & 23.46 & \textbf{32.60} & 9.38 \\ 
    \rowcolor{gray!20}WD & \textbf{20.91} & 27.57 & \textbf{18.77} & \textbf{16.03} & \textbf{5.72} & \textbf{21.48} & \textbf{15.20} & \textbf{23.47} & 32.48 & \textbf{10.09}   \\
    \bottomrule
  \end{tabular}
  \caption{Comparison of Different Box Regression Losses: Corner Chamfer Distance Loss (CCD), L1 Loss, Permutation Corner Distance Loss (PCD), Wasserstein Distance Loss (WD).}
  \label{tab:ablation loss}
\end{table*}

\begin{table*}
  \centering
  \begin{tabular}{@{}l|c|cc|cc|ccc@{}}
    \toprule
    Number of Desc & Overall & Easy & Hard & View-dep & View-indep  & ScanNet & 3RScan & MP3D\\
    \midrule
    1 & 53.10 & 53.16  & \textbf{52.50}  & 55.09 & 51.98 & 59.60 & 52.22 & \textbf{40.73}\\
    10 & 49.05 & 49.50  & 44.04 & 50.71 &  48.12  &   54.90 & 50.87 & 35.55 \\
    \rowcolor{gray!20}random 1 to 10& \textbf{54.66} & \textbf{55.07} & 50.12 & \textbf{55.78}  & \textbf{54.03}  &  \textbf{61.23} & \textbf{55.41} & 39.36\\
    \bottomrule
  \end{tabular}
  \caption{Ablation Study on Number of Descriptions during Training for Visual Grounding 3D.}
  \label{tab:ablation grounding}
\end{table*}

\subsection{Benchmark}
We use the EmbodiedScan benchmark to validate the effectiveness of BIP3D.
%
EmbodiedScan is a 3D indoor dataset comprising 4,633 high-quality scans from ScanNet, 3RScan, and Matterport3D (MP3D), with 1,513, 1,335, and 1,785 scans from each source, respectively.
The training, validation, and testing sets contain 3,113, 817, and 703 scans, respectively. The data from ScanNet, 3RScan, and Matterport3D all include RGB-D images, but there are significant differences in camera types, which impose higher requirements on model performance.

We use 3D Intersection over Union (IoU)-based Average Precision (AP) with a threshold of 0.25, AP$_{3D}$@0.25, as the evaluation metric. For the 3D detection task, to provide a detailed performance analysis, we adopt three sets of sub-metrics:
\textbf{(1) Category generalization:} To assess generalization across object categories, we follow EmbodiedScan~\cite{wang2024embodiedscan}, divide objects to be detected into head, common, and tail categories and compute metrics for each.
\textbf{(2) Performance on small objects:} To highlight the advantages of an image-centric approach, we categorize objects by volume into small, medium, and large parts and compute metrics for each volume part.
\textbf{(3) Scene and sensor generalization:} To further evaluate the model's generalization across different scenes and sensor types, we calculate metrics for different subsets.
For the 3D grounding task, we adopt two sets of sub-metrics from EmbodiedScan~\cite{wang2024embodiedscan}:
\textbf{(1) Difficulty:} A sample is considered hard if the number of distracting targets exceeds three.
\textbf{(2) View dependency:} If the text contains directional description such as “front/back” or “left/right,” the sample is deemed view-dependent.


\subsection{Implementation Details}
We use Swin-Transformer-Tiny as the image backbone, BERT-Base as the text encoder, and a mini-ResNet34 (with channel numbers reduced to one-quarter of the standard model) as the depth backbone. Both the feature enhancer and the decoder consist of six transformer layers. We load GroundingDINO-tiny as our pretrained model. Table~\ref{tab:parameters} demonstrates the comparison of the number of parameters between our model and the point-centric grounding model, EmbodiedScan. It is evident that the parameters of our 3D encoder is significantly lower than that of point-centric model.

We implement view-dependent queries, with 50 queries assigned per view. These 50 queries are obtained by projecting the ground truth 3D bounding boxes from the training set into the camera coordinate system and then clustering them by K-Means. Similar to GroundingDINO, we use sub-sentence level representations to encode text prompts.

During the training phase, we randomly sample 18 images as input, while during testing, we select 50 key frames at fixed intervals as input. For 3D detection, we extract 128 class names as text for each training sample, whereas during testing, all 284 class names are used as textual input. For 3D grounding, we randomly select between 1 and 10 textual descriptions per training sample, and during testing, only one text prompt is used per instance.
All models are trained using 8 NVIDIA 4090 GPUs with 24GB of memory, employing the AdamW optimizer. More parameters are detailed in the appendix.

\subsection{Main Results}
\textbf{Detection 3D.} We selected several representative methods of different types for comparison: (1) the point model VoteNet~\cite{votenet}, (2) the RGB-only model ImVoxelNet~\cite{rukhovich2022imvoxelnet}, (3) the point-sparse-voxel model FCAF3D~\cite{rukhovich2022fcaf3d}, and (4) the multi-modal fusion model EmbodiedScan~\cite{wang2024embodiedscan}. Table~\ref{tab:detection3d-main} presents the metrics of each method, showing that BIP3D significantly outperforms existing methods, with an AP$_{3D}$@0.25 on the overall dataset that is 5.69\% higher than EmbodiedScan. Notably, thanks to the 2D pre-trained model, BIP3D exhibits excellent category generalization performance, achieving an AP of 16.03\% on tail categories, which far exceeds EmbodiedScan’s 9.48\%. Moreover, due to the dense nature of image features, BIP3D also achieves superior performance on small objects, with an AP of 5.72\%, compared to a maximum of 3.28\% for other methods. When BIP3D uses only RGB as input, the AP decreases by 3.51\%, but it still surpasses all existing methods. The input of depth primarily affects the localization precision, which is notably reflected in a more pronounced decrease in AP for small objects, while there is no significant change for large objects.

\noindent \textbf{Visual Grounding 3D.}
The comparison of our method with others on the 3D visual grounding benchmark is shown in Table~\ref{tab:grounding-main}. First, on the validation dataset, our BIP3D overall AP surpasses EmbodiedScan by 15.25\%. Furthermore, it can be observed that our method demonstrates better robustness; the performance on hard samples decreases by only 4.95\% compared to easy samples, while for EmbodiedScan, the decrease is 8.67\%. On the test dataset, without model ensemble, BIP3D achieves an AP of 57.05\%, which is 17.38\% higher than that of EmbodiedScan; with model ensemble, the AP of BIP3D further improves to 62.08\%, surpassing the state-of-the-art solution DenseG by 2.49\%.

\subsection{Ablation Studies and Analysis}
\noindent \textbf{Pretraining.} To demonstrate the significant role of 2D pretraining for 3D perception tasks, we conducted ablation studies on both the point-centric model EmbodiedScan and the image-centric model BIP3D. Firstly, we found that for EmbodiedScan, initializing with GroundingDINO weights brought only a marginal improvement of 1.11\%, with a mere 0.38\% increase for tail categories, indicating minimal effect. Conversely, for BIP3D, the use of GroundingDINO weights resulted in substantial improvements of 5.66\% and 5.99\% for the RGB-only and RGB-D models, respectively. This suggests that effective initialization can significantly enhance the performance of 3D detection, highlighting the importance for 3D models to fully leverage 2D foundation models.

\noindent \textbf{Camera Intrinsic Standardization.} Table~\ref{tab:ablation cis} demonstrates the effect of Camera Intrinsic Standardization (CIS) on 3D detection performance. It can be observed that CIS brings about a performance improvement of 0.7-1.3\% for both RGB-only and RGB-D models. When trained exclusively on Scannet, incorporating CIS results in a significant boost in performance on unseen camera data. However, due to notable differences in scenes and categories across different evaluation datasets, other aspects of generalization continue to affect the model's transferability.

\noindent \textbf{Box Regression Loss.}
We compared several bounding box regression losses. It is evident that when using L1 distance directly, the AP$_{3D}$@0.25 only reaches 17.79\%, which is due to the inability of L1 distance to handle the ambiguity in box orientation, leading to incorrect optimization directions. The corner chamfer distance can avoid such orientation ambiguity; however, corner chamfer distance loss training on BIP3D is unstable and difficult to converge. Both permutation corner distance loss and Wasserstein distance loss are orientation-agnostic, avoiding orientation ambiguity and enabling the model to converge stably, achieving better performance. Specific results are shown in the Table~\ref{tab:ablation loss}. Detail about permutation corner distance loss refers to appendix.

\noindent \textbf{Number of Description.} During training for visual grounding, using only one text description per sample results in low training efficiency. However, directly employing multiple descriptions for training can introduce a domain gap during testing, leading to degraded performance. Therefore, we opted to randomly select between 1 and 10 text descriptions for training. Table~\ref{tab:ablation grounding} demonstrates the impact of the number of descriptions on model performance. It is evident that our random selection strategy improves AP by 1.56\% compared to using a single description, and by 5.61\% compared to using a fixed set of 10 descriptions.

\section{Conclusion and Future Works}
In this work, we propose an image-centric 3D perception model, BIP3D. It overcomes the limitations of point clouds and effectively leverages the capabilities of 2D foundation models to achieve significant improvements in 3D perception performance. BIP3D supports multi-view images, depth maps, and text as inputs, enabling it to perform 3D object detection and 3D visual grounding. We demonstrate the superiority of BIP3D on the EmbodiedScan benchmark.

BIP3D still has considerable room for exploration, and several future works are outlined here:
(1) Further optimizing the network architecture and training schemes to achieve even better perception performance.
(2) Applying BIP3D to dynamic scenes to achieve joint detection and tracking.
(3) Incorporating more perception tasks, such as instance segmentation, occupancy, and grasp pose estimation.
(4) Under the integrated network framework of BIP3D, the decoder can be improved to support higher-level tasks like visual question answering and planning.

\newpage
{
    \small
    \bibliographystyle{ieeenat_fullname}
    \bibliography{main}
}

\appendix
\clearpage
\setcounter{page}{1}
\renewcommand{\thetable}{A.\arabic{table}}
\renewcommand{\thefigure}{A.\arabic{figure}}
\setcounter{table}{0}
\setcounter{figure}{0}
\maketitlesupplementary

\section{Ablation on 3D Position Embedding}
Position embedding in the spatial enhancer is a crucial component of BIP3D, serving to bridge the gap between 2D image features and 3D space. Table~\ref{tab:3DPE} demonstrates the impact of 3D PE on detection performance. When the 3D PE is removed, the overall AP decreases by 3.09\%.

\begin{table}[h]
  \centering
  \begin{tabular}{@{}c|c|ccc@{}}
    \toprule
    3D PE & Overall & Head & Common & Tail \\
    \midrule
    & 17.82 & 23.63 & 14.34 & 15.39  \\
    \checkmark  & \textbf{20.91} & \textbf{27.57} & \textbf{18.77} & \textbf{16.03} \\
    \bottomrule
  \end{tabular}
  \caption{Ablation Results of 3D PE.}
  \label{tab:3DPE}
\end{table}

\begin{figure*}
    \centering
    \vspace{-0.5cm}
    \includegraphics[width=0.9\linewidth]{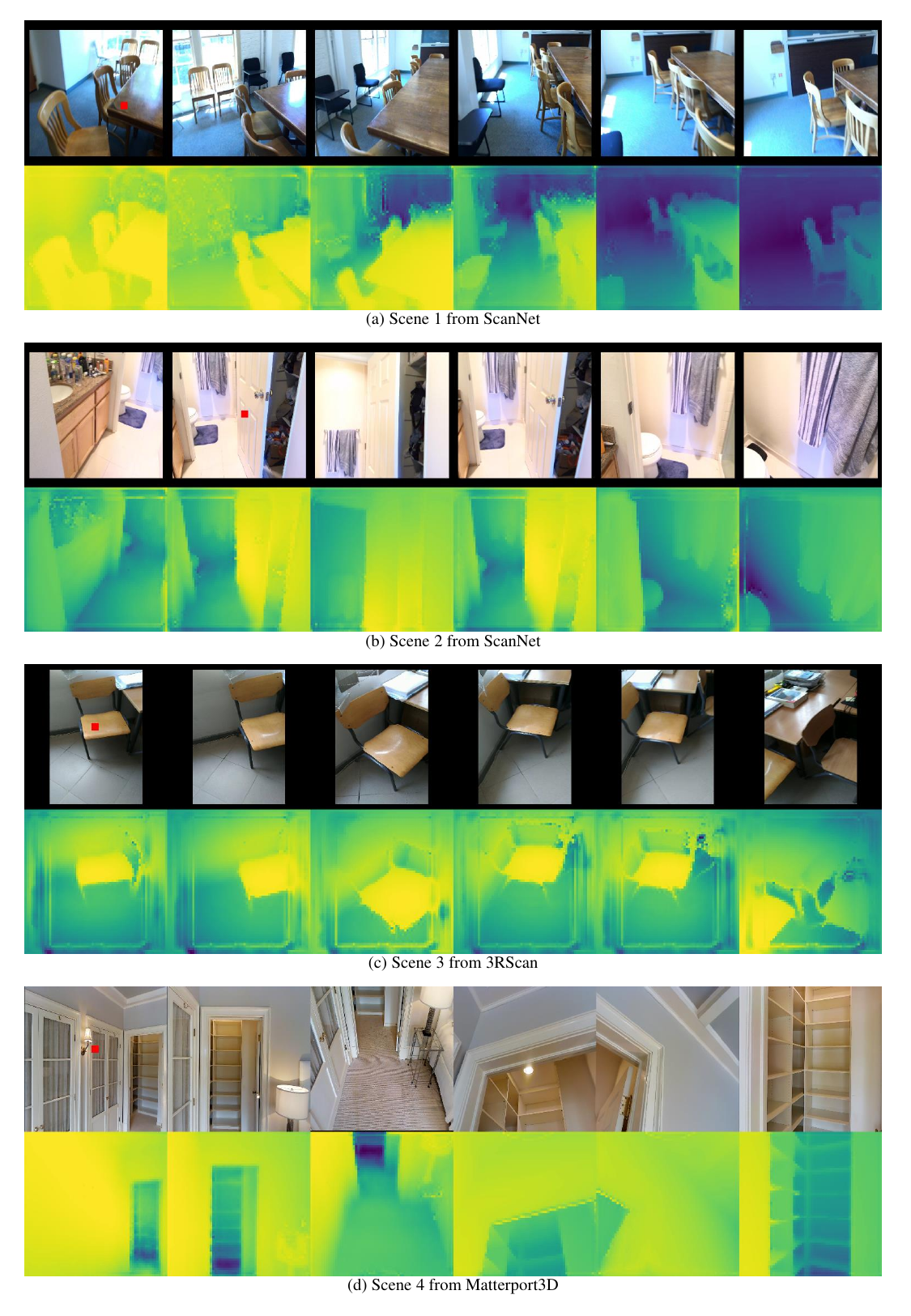}
     \vspace{-0.5cm}
    \caption{Visualization of the Correlations of Position Embeddings. The red boxes on the images indicate the selected target location, while the heatmaps represent the cosine similarity between all position embeddings and the position embedding of the target location.}
    \label{fig:ptsembedvis}
\end{figure*}

To more intuitively illustrate that the spatial enhancer achieves spatial modeling, we visualize the correlations between 3D position embeddings $\bm{\mathrm{IPE}}$. As shown in Figure~\ref{fig:ptsembedvis}, it can be observed that embedding correlations exhibit a positive relationship with their 3D positions.

\section{Inference Efficiency}
\begin{figure}[h]
    \centering
    \includegraphics[width=0.99\linewidth]{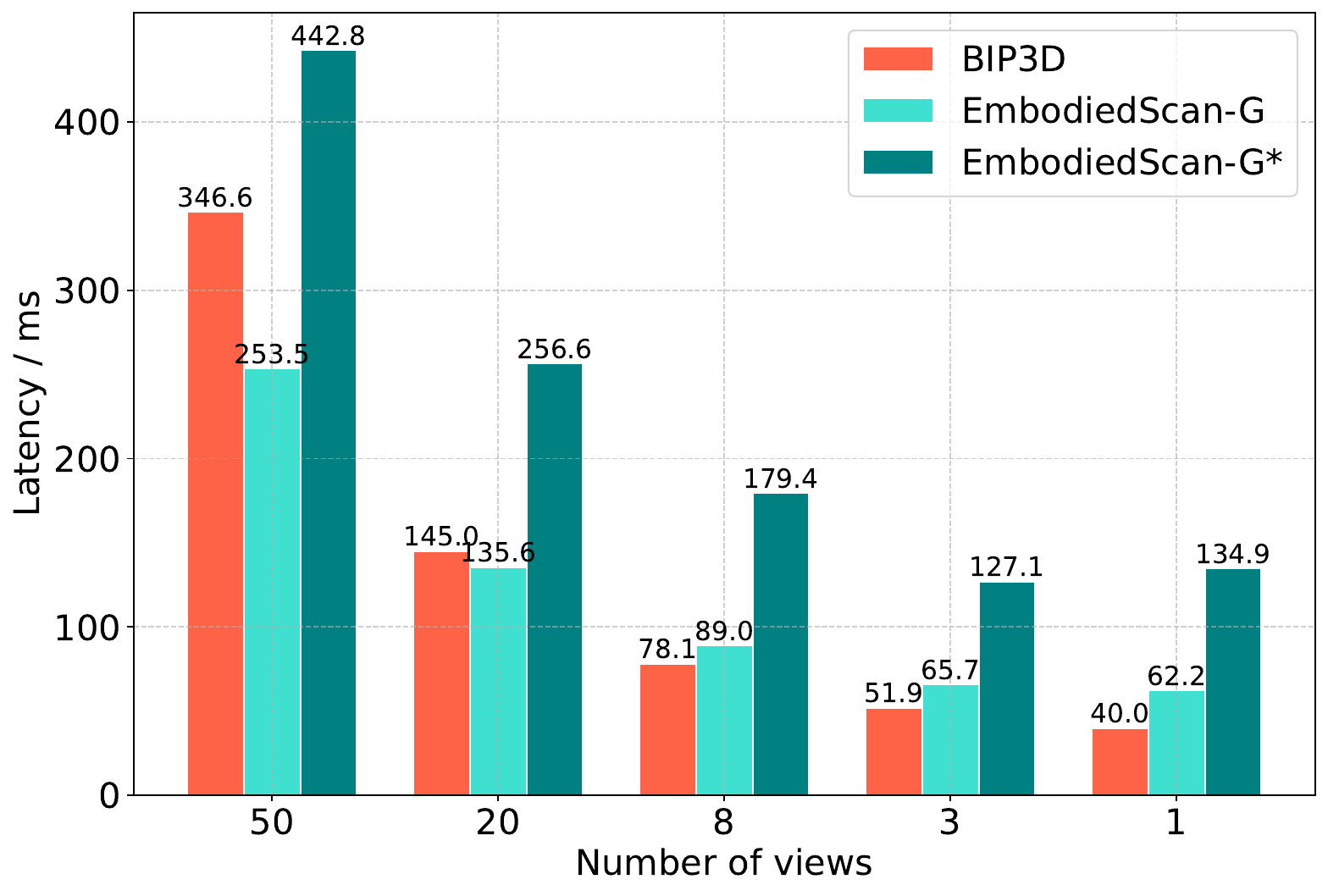}
    \vspace{-0.3cm}
    \caption{Latency Comparison, where `*' indicates the inclusion of point cloud preprocessing time, encompassing multi-view aggregation and down-sampling.}
    \label{fig:latency}
\end{figure}

We compared the inference speeds of BIP3D and EmbodiedScan on a 4090 GPU, as shown in Figure~\ref{fig:latency}. When considering the point cloud preprocessing time, BIP3D consistently exhibits lower latency than EmbodiedScan, with a more pronounced advantage when the number of views is small. When focusing solely on the neural networks, BIP3D's inference speed is slower at a higher number of views; however, when the number of views is reduced to below 8, BIP3D still maintains an efficiency advantage.

\section{Scale-up}
To test the impact of increasing model parameters on perception performance, we replaced the backbone with Swin-Transformer-Base. As shown in Table~\ref{tab:scaleup}, compared to Swin-Tiny, the overall AP improved by 1.21\%, with a significant increase of 2.74\% in long-tail categories. It is worth noting that GroundingDINO-Base showed an improvement from 58.1\% to 59.7\% over GroundingDINO-Tiny on COCO benckmark.
To further enhance model performance, we incorporated additional training data from the ARKitScene dataset. This resulted in an additional 1.47\% improvement. These results highlight the positive impact of both scaling up the model size and enriching the training dataset on improving detection accuracy.

\begin{table}[h]
  \centering
  \begin{tabular}{@{}l|c|c|ccc@{}}
    \toprule
    Backbone & ARKit & Overall & Head & Common & Tail \\
    \midrule
    swin-tiny & & 20.91 & 27.57 & 18.77 & 16.03 \\
    swin-base & & 22.12 & 28.63 & 18.77 & 18.77\\
    swin-base & \checkmark & \textbf{23.59} & \textbf{30.20} & \textbf{19.59}  & \textbf{20.88}\\
    \bottomrule
  \end{tabular}
  \caption{Results of Scale-up Experiments.}
  \label{tab:scaleup}
\end{table}

\section{Detail of Camera Intrinsic Standardization}
\begin{figure}[h]
    \centering
    \includegraphics[width=0.99\linewidth]{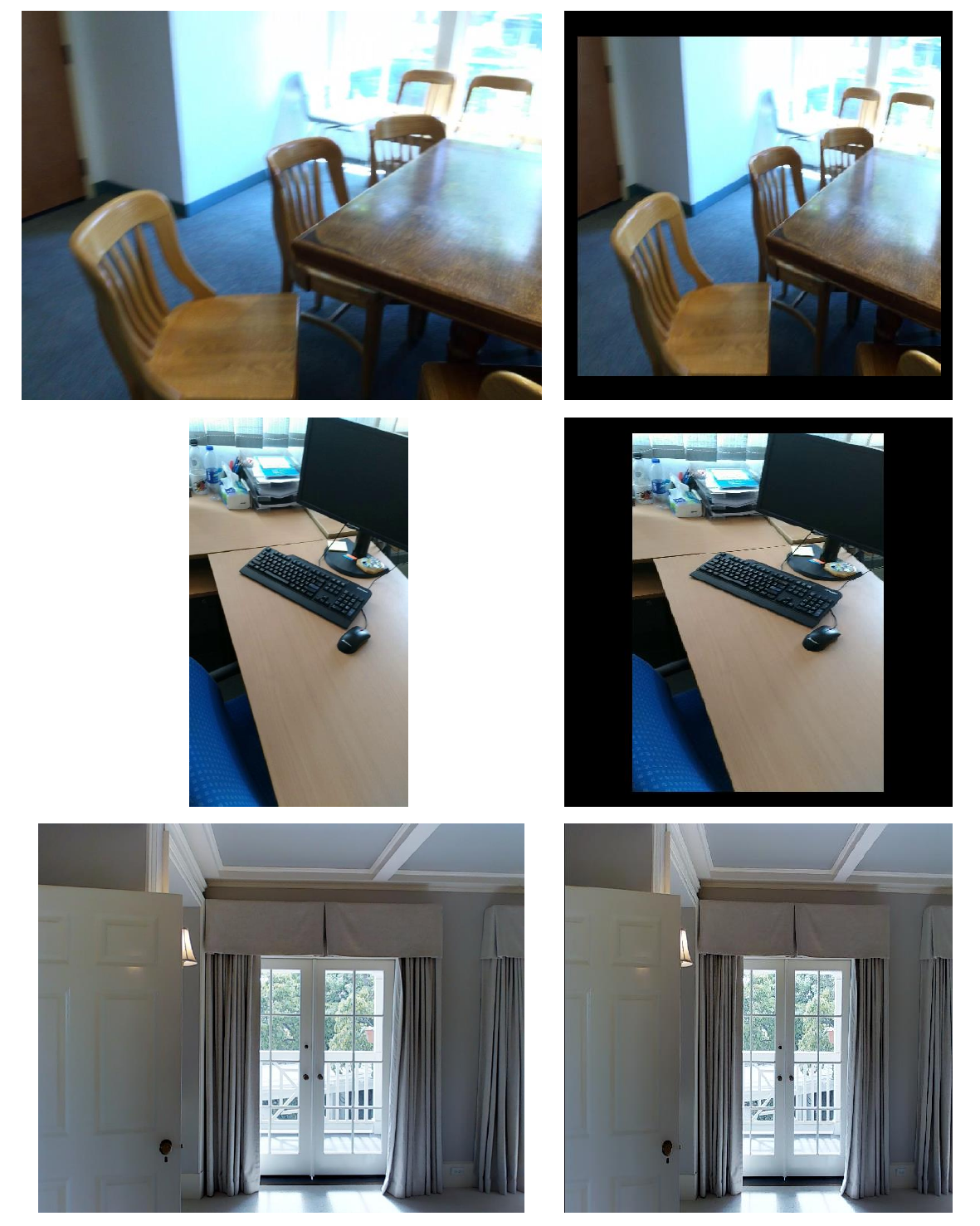}
    \vspace{-0.3cm}
    \caption{Images Comparison Before and After Camera Intrinsic Standardization. Left: Original, Right: Standardized.}
    \label{fig:cissample}
\end{figure}

The parameters of standardized intrinsic are derived from the mean of the training set. Given that we use undistorted pinhole cameras, the parameters include $[focal_u, focal_v, center_u, center_v]$, which are set to $[432.579,539.857,256,256]$. Intrinsic standardization may introduce issues such as pixel loss and zero padding, as shown in Figure~\ref{fig:cissample}.

\section{Model Ensemble}

For the model ensemble experiment listed in Table 3 of the main text, we employ five models. Two of these models are trained on the entire dataset, utilizing permutation corner distance loss and Wasserstein distance loss, respectively. The remaining three models are trained on distinct data subsets: ScanNet, 3RScan, and Matterport3D. The strategy for model ensemble is 3D NMS with 0.4 IoU threshold.

\section{Permutation Corner Distance Loss}
\label{sec:PCL}
\begin{figure}[h]
    \centering
    \includegraphics[width=0.99\linewidth]{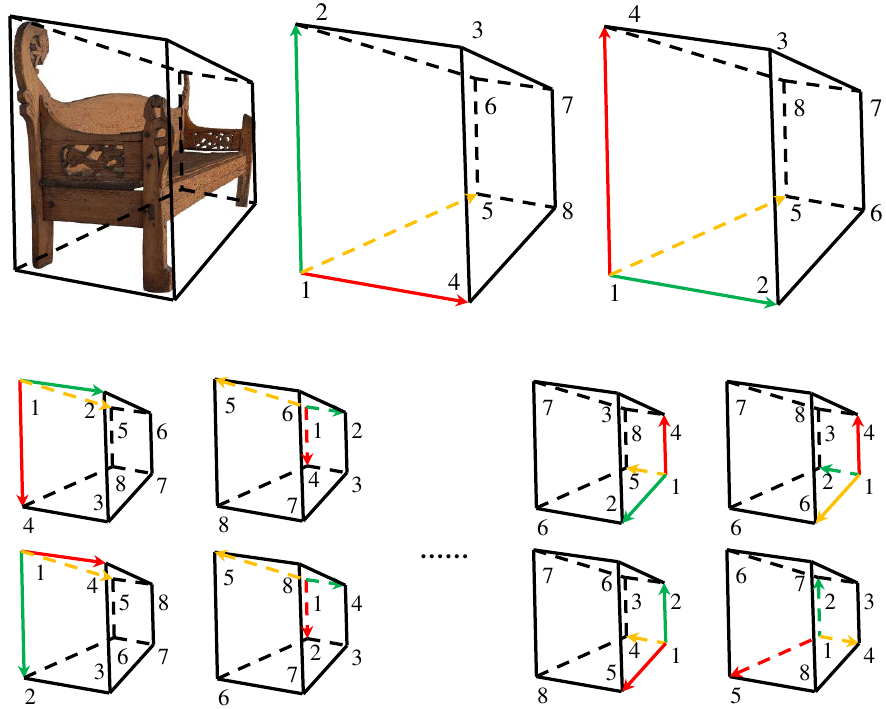}
    \caption{The 3D Bounding Box Corners Permutations. For the same bounding box, there are a total of 48 different corner point permutation; the corner point order is indicated by numbers, with red, yellow, and green representing width, length, and height, respectively.}
    \label{fig:permutation_corners}
\end{figure}

For a single 3D bounding box, there are 48 possible permutations of its 8 corner points, denoted as $\mathcal{A}$, as shown in Figure~\ref{fig:permutation_corners}. Different permutations correspond to different 
$[w,l,h,roll,pitch,yaw]$ values. Therefore, using $||\bm{\mathrm{B}}_{pred}-\bm{\mathrm{B}}_{gt}||$ directly as the loss function would result in incorrect gradients. We propose a permutation corner loss defined as:
\[
L_{box}=\min_{1\leq i\leq  48}\left[||\mathcal{A}(pred)_1-\mathcal{A}(gt)_i||_2\right]
\]

\section{Model Prediction Visualization}
\begin{figure*}
    \centering
    \includegraphics[width=0.96\linewidth]{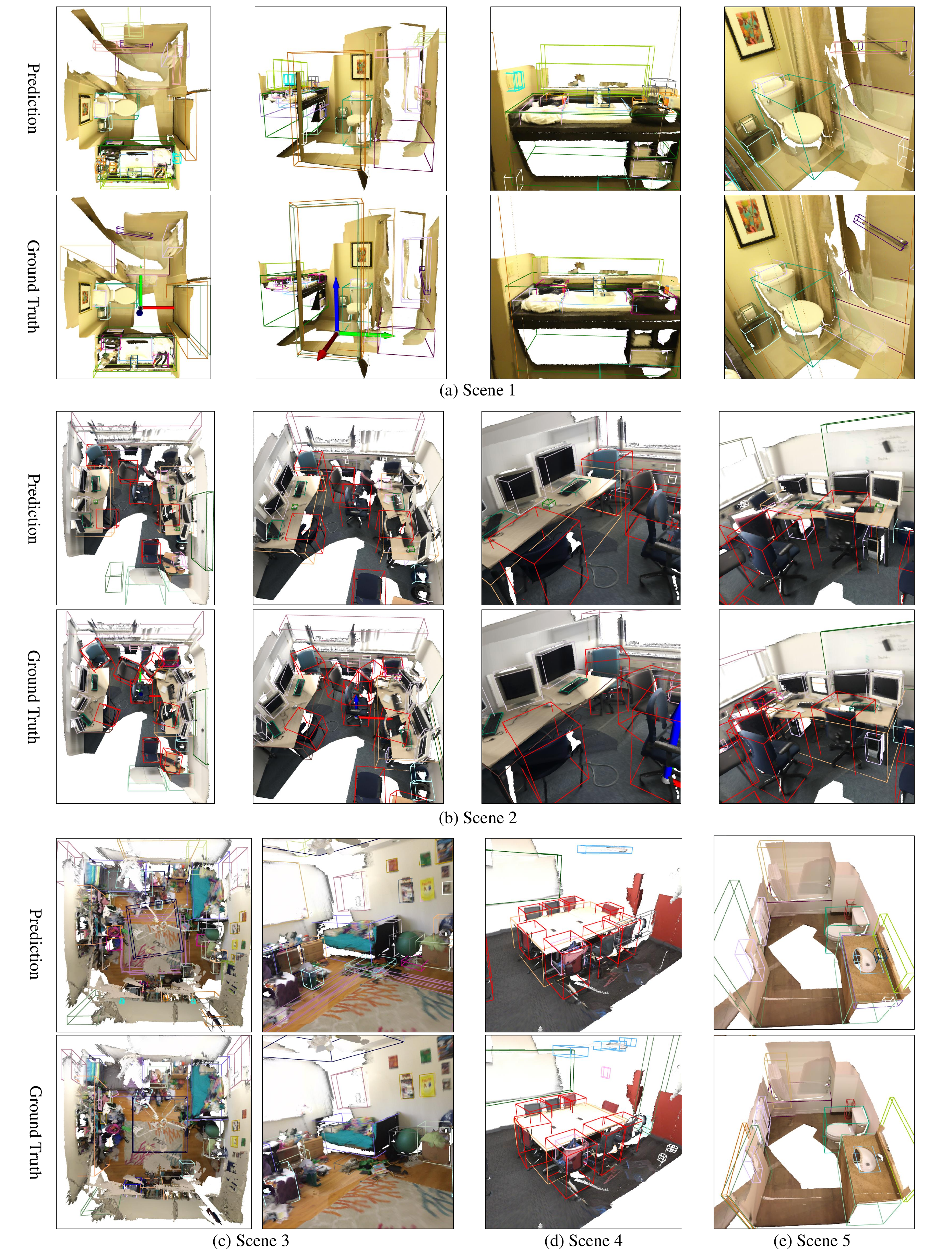}
    \caption{Visualization of 3D Detection Results. The color of the boxes indicates the category.}
    \label{fig:detvis}
\end{figure*}

\begin{figure*}
    \centering
    \includegraphics[width=0.85\linewidth]{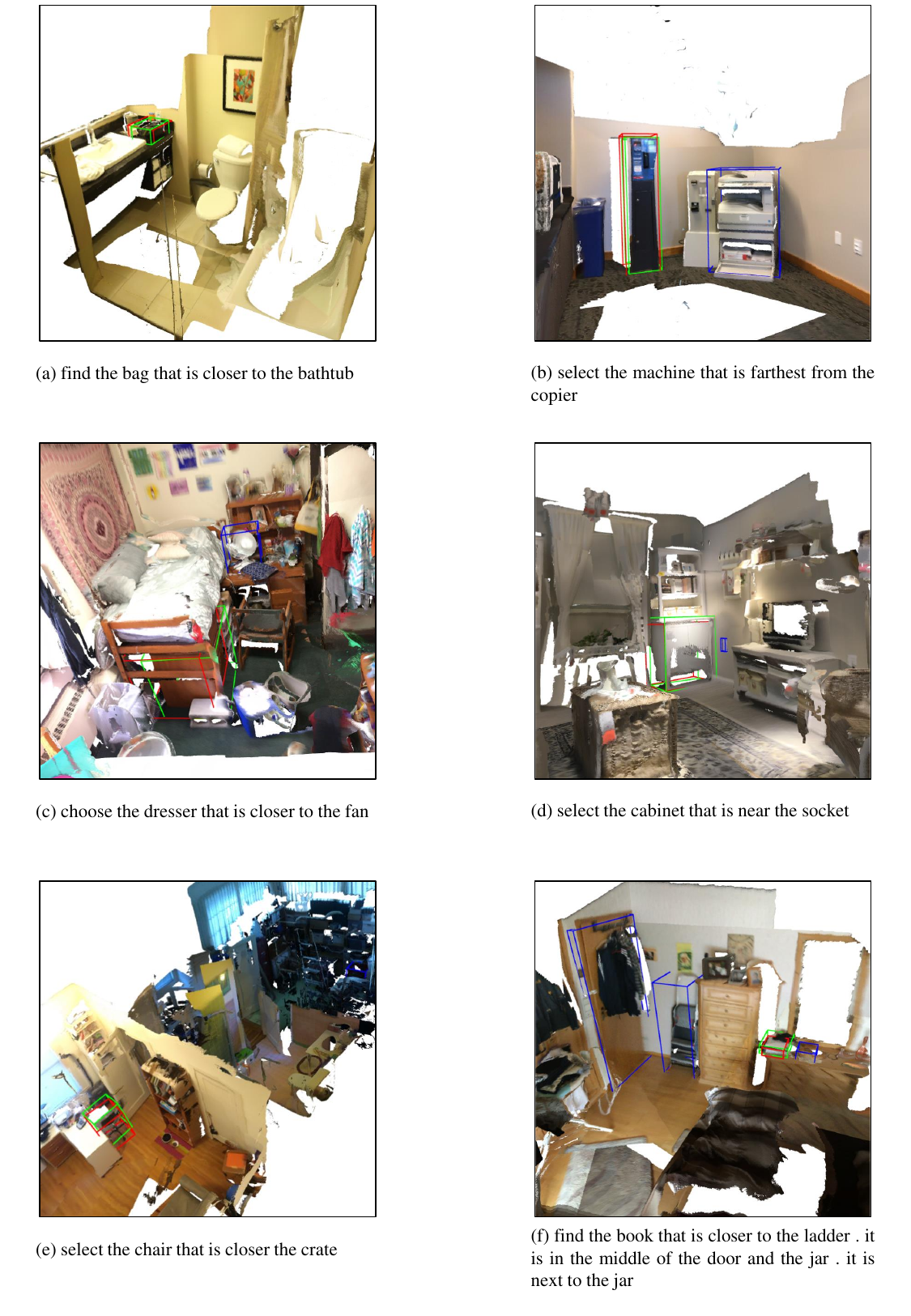}
    \vspace{-0.2cm}
    \caption{Visualization of 3D Visual Grounding. Green boxes represent the ground truth, red boxes represent the predictions, and blue boxes represent reference objects, such as the `bathtub' in `find the bag that is closer to the bathtub'.}
    \label{fig:groundingvis}
\end{figure*}

Figure~\ref{fig:detvis} visualizes the 3D detection results of the model, demonstrating that BIP3D can effectively handle a variety of complex indoor scenarios. Even for some objects that are unannotated, BIP3D is capable of detection, which provides feasibility for enhancing model performance through the use of semi-supervised learning in the future.  Figure~\ref{fig:groundingvis} visualizes the 3D grounding results, illustrating the model's capability to identify and locate the specific target designated by the text among multiple objects of the same class.

\section{Algorithm Setting}
Table~\ref{tab:config} lists more detailed model configurations and training parameters.
Additionally, we employ two types of data augmentation during training: 1) applying a random grid mask to the depth map, and 2) performing random cropping on both the images and depth maps.

\begin{table}[h]
    \centering
    \begin{tabular}{c|c}
        \toprule
         Config & Setting \\
         \midrule
         image backbone & swin-transformer-tiny \\
         image neck & channel mapper \\
         depth backbone & mini-ResNet34 \\
         depth neck & channel mapper \\
         text encoder & BERT-base \\
         embed dims & 256 \\
         feature levels & 4 \\
         key points & 7 fixed and 9 learnable \\
         feat enhancer layers & 6 \\
         decoder layers & 6\\
         anchor per view & 50 \\
         max depth $D$ & 10 \\
         num of points $K$ & 64 \\
         
         optimizer & AdamW \\
         base lr & 2e-4 \\
         image backbone lr & 2e-5 \\
         text encoder lr & 1e-5 \\
         detection epochs & 24 \\
         grounding epochs & 2 \\
         batch size & 8 \\
         weight decay & 5e-4 \\
         drop path rate & 0.2 \\
         $\lambda_1$ & 1.0 \\
         $\lambda_2$ & 0.8 \\
         $\lambda_3$ & 1.0 \\
         dn queries & 100 \\
         training views & 18 \\
         test views & 50 \\
         \bottomrule
    \end{tabular}
    \caption{Model Configurations and Training Parameters.}
    \label{tab:config}
\end{table}

\vspace{-0.5cm}

\vspace{-0.5cm}

\end{document}